\documentclass{llncs}

\usepackage{amsmath,amsfonts,amssymb,amsthm}
\usepackage{verbatim}
\usepackage{color}
\usepackage{epsfig}
\usepackage{array}
\usepackage{algorithm}
\usepackage{cite}
\usepackage{tabu}

\begin{document}

\title{Ischemic Stroke Lesion Segmentation in CT Perfusion Scans using Pyramid Pooling and Focal Loss}

\titlerunning{Stroke Lesion Segmentation in Perfusion Images}  

\author{S. Mazdak Abulnaga\inst{1,2} \and Jonathan Rubin\inst{2} }

\authorrunning{S.M. Abulnaga \etal} 

\institute{Computer Science and Artificial Intelligence Lab, MIT, Cambridge, MA, USA\\
\and Philips Research North America, Cambridge, MA, USA\\
\email{abulnaga@mit.edu, jonathan.rubin@philips.com}}

\maketitle
\begin{abstract}
We present a fully convolutional neural network for segmenting ischemic stroke lesions in CT perfusion images for the ISLES 2018 challenge. Treatment of stroke is time sensitive and current standards for lesion identification require manual segmentation, a time consuming and challenging process. Automatic segmentation methods present the possibility of accurately identifying lesions and improving treatment planning. Our model is based on the PSPNet, a network architecture that makes use of pyramid pooling to provide global and local contextual information. To learn the varying shapes of the lesions, we train our network using focal loss, a loss function designed for the network to focus on learning the more difficult samples. We compare our model to networks trained using the U-Net and V-Net architectures. Our approach demonstrates effective performance in lesion segmentation and ranked among the top performers at the challenge conclusion.

\end{abstract}
\section{Introduction}
\vspace{-0.1in}

We present a model for segmenting stroke lesions in CT perfusion (CTP) data for the 2018 ischemic stroke lesion segmentation (ISLES) challenge. Ischemic stroke is caused by an obstruction of blood supply to the brain. Treatment of stoke is time sensitive, requiring tissue reperfusion within less than 4-6 hours of stroke onset~\cite{maier2017isles}. Current standards for evaluating stroke requires manual segmentation in MRI or CT images~\cite{maier2017isles,GILLEBERT2014CTanalysis,biesbroek2013diagnostic}, a challenging and time consuming task, due to the changing appearance of lesions over time and their presence in various locations in the brain~\cite{maier2017isles,winzeck2018isles}. There is a growing need for automatic segmentation methods to accurately identify lesions and to help plan effective treatment.

The 2018 ISLES challenge is the first to use CTP data. Currently, MR with diffusion-weighted imaging (DWI) is considered the most accurate and earliest at detecting acute stroke~\cite{GILLEBERT2014CTanalysis,biesbroek2013diagnostic}. CTP however is advantageous in cost, speed, and availability in acute care units~\cite{GILLEBERT2014CTanalysis}. Furthermore, CTP is emerging as an effective means to detect the infarct (irreversible) core with high sensitivity and specificity~\cite{GILLEBERT2014CTanalysis,biesbroek2013diagnostic}. Detection relies on quantitative evaluation metrics derived from the CTP data. For example, a drop in cerebral blood flow (CBF) is a result of an occlusion of blood supply~\cite{campbell2011cbf}. In this work, we use the CT image, CBF, cerebral blood volume (CBV), time to peak (TTP) and mean transit time (MTT) of the contrast agent injection as signals to identify the infarct core. 

\begin{figure}
\centering
\includegraphics[width=0.9\columnwidth]{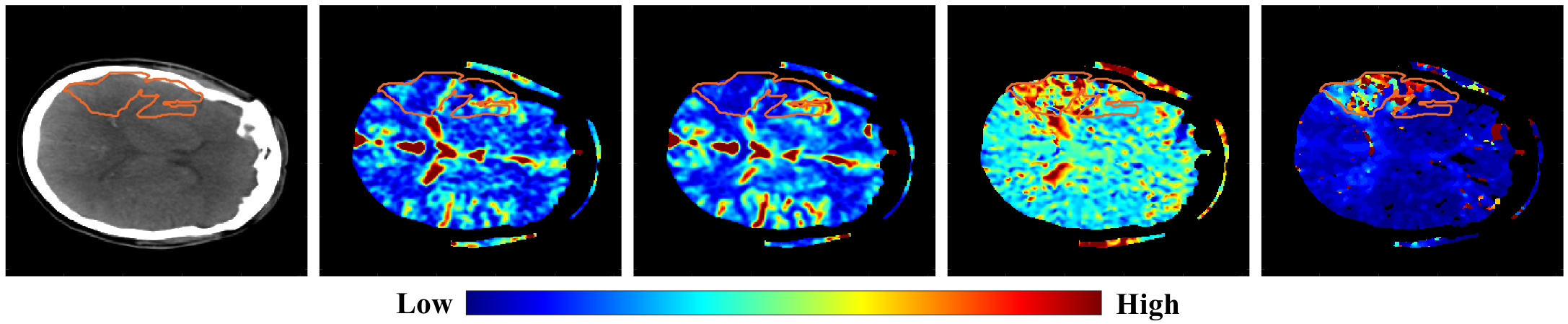}
\caption{Example CTP data from one subject in the study, with stroke lesion segmentation overlaid in orange. From left to right: CT, CBV, CBF, MTT, Tmax images. Each image had different units of measure, so we demonstrate for visualization only.}
\label{fig:intro}
\end{figure}
\vspace{-0.1in}
\section{Related Work}
\vspace{-0.1in}
Deep learning approaches that utilize fully convolutional neural network (CNN) architectures~\cite{ronneberger2015unet,milletari2016vnet} have become the de facto standard for semantic segmentation tasks in 2D and 3D medical imaging~\cite{litjens2017survey}. The ISLES challenge was established to fairly compare approaches in stroke lesion segmentation and characterization~\cite{maier2017isles,winzeck2018isles}, resulting in the development of effective CNN models for this task~\cite{kamnitsas2017efficient,chen2017fully,guerrero2018white}. All previous challenges focused on multispectral MRI data~\cite{maier2017isles,winzeck2018isles}. Many early approaches focused on analyzing patches, in part due to memory issues. A top performer of the 2015 challenge developed a 3D patch-based CNN architecture that used two parallel pathways, allowing the network to process patches at different scales~\cite{kamnitsas2017efficient,maier2017isles}. More recent efforts in the 2016 and 2017 challenges investigated extensions to dual pathway 3D networks~\cite{robben2017dual} and ensembles of multi-scale networks~\cite{choi2016ensemble}. In the 2017 challenge, the authors of~\cite{choi2016ensemble} investigated an additional network based on \emph{pyramid scene parsing}~\cite{zhao2016pyramid}. The models we present in this work also make use of pyramid pooling as in~\cite{zhao2016pyramid}, though we focus on CTP data. There have been few works exploring the automatic segmentation of stroke lesions using CT data~\cite{REKIK2012ctreview}, and to the best of our knowledge, none have made use of deep neural networks. Many previous methods have relied on histogram-based classifiers~\cite{REKIK2012ctreview}, or make use of statistical comparisons for lesion detection~\cite{GILLEBERT2014CTanalysis}. 

Many other medical image segmentation tasks have benefited from using deep learning-based classifiers, for example in pancreas segmentation using CT~\cite{oktay2018attention}, prostate segmentation using MRI~\cite{milletari2016vnet}, and multi-organ segmentation in whole-body CT~\cite{brosch2018foveal}. Finally, it is also worth mentioning that public datasets and challenges, such as The Pascal Visual Object Classes Challenge (VOC)~\cite{everingham2010pascal}, have resulted in significant improvements in natural image semantic segmentation using CNNs. Advances in natural images have also informed medical image segmentation tasks. In this work, we utilize a modified 2D fully convolutional architecture that was pre-trained on natural images from the PASCAL VOC dataset. We build upon models that have demonstrated effectiveness in both medical and natural image segmentation tasks. Furthermore, we are one of the first to develop a deep network to automatically segment stroke lesions in CTP data, and demonstrate strong performance compared to other challenge participants.

\section{Dataset}
\label{s:dataset}
\vspace{-0.1in}
The ISLES challenge data included the CT scan, the CTP source data, and the CBF, CBV, MTT, and Tmax derived perfusion maps, though we did not use the CTP source data. Images were acquired within 8 hours of stroke onset. An MRI DWI was then acquired within 3 hours after the CTP scan. The infarct core lesions were manually drawn using the corresponding MRI DWI scans. CTP scans were acquired as slabs covering sparse areas (5mm axial spacing) with stroke lesion in the brain. As a result, the scans had varying depth in the axial dimension, ranging from 2 to 22 slices. Each slice was a $256 \times 256$ image. Furthermore, some patients had two non-overlapping or partially-overlapping slabs covering regions within the brain. The training set contained 63 subjects and 94 scans, and the test set contained 40 subjects with 62 scans.
\section{Methods}
\label{sec:methods}
\vspace{-0.1in}

Fully convolutional neural network architectures were trained to predict ischemic stroke lesion masks. We constructed both 2D and 3D CNN models, but found stronger performance in 2D per-slice models given the variable and limited number of axial slices in the scans. 
The input to the network was a multi-channel 2D image created by stacking a CT slice together with its four corresponding perfusion map slices (Tmax, CBF, CBV, MTT). Cross entropy and focal loss~\cite{lin2018focal} were evaluated as loss functions. We developed models based on the pyramid scene parsing network (PSPNet), ~\cite{zhao2016pyramid}, the U-Net (2D and 3D)~\cite{ronneberger2015unet,cciccek20163dunet}, and the 3D V-Net~\cite{milletari2016vnet} architectures. Our final model is based on the PSPNet with focal loss. The PSPNet employs \emph{pyramid pooling} (explained below) within a fully convolutional neural network.

\vspace{-0.1in}
\subsection{Data Augmentation}
Data augmentation was used to artificially increase the size of our limited training set. We augmented the images sagitally and coronally to reflect likely variations in appearance of the brain and stroke lesions. The augmentation would randomly rotate the images by $[-10^\circ,10^\circ]$, translate by $[-10\%,10\%]$ of the image size, flip, and scale by a factor of $[0.9, 1.1]$. The sampling was done uniformly and the order of these operations was chosen randomly.

\vspace{-0.1in}
\subsection{Pyramid Scene Parsing Network}
The pyramid scene parsing architecture was chosen as it achieves state-of-the-art performance on segmentation tasks in natural images. In particular, it achieved first place at the 2016 ImageNet scene parsing challenge \cite{russakovsky2015imagenet}.

PSPNet combines a ResNet-based~\cite{he2016deep} fully convolutional neural network architecture~\cite{long2015fully} together with dilated convolutions~\cite{chen2014semantic,yu2015multi}. Further, the PSPNet introduces a pyramid pooling module that performs region-based context aggregation. The pyramid pooling module is designed to capture global information about an input image from different regions of a network's receptive field and at various scales. To do so, pooling kernels of varying sizes and strides are applied to a network's final feature map layer.
We adopt the same dimensions of the four level pyramid pooling module as described in~\cite{zhao2016pyramid}.

\subsubsection{Pyramid Pooling Module}
\begin{figure}
\includegraphics[width=1.0\columnwidth]{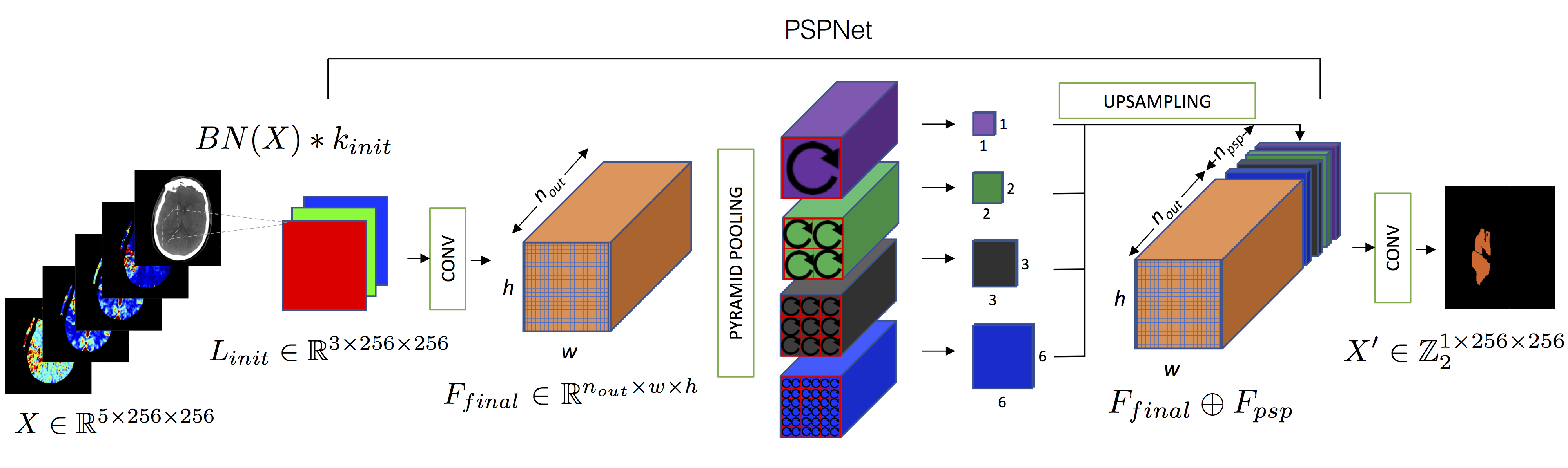}
\caption{Architecture of a fully convolutional neural network with a pyramid pooling module for segmenting ischemic stroke lesions. Circular arrows represent average pooling operations. Input to the model is a stack of $256\times256$ multi-modal CT perfusion maps. The model outputs a $\mathbb{Z}_2 = \{0,1\}$, single channel $256\times256$ prediction mask.}
\label{fig:psp}
\end{figure}

Fig. \ref{fig:psp} illustrates the module graphically. Consider the last convolutional layer of a network, $L_{final}$ that consists of $n_{out}$ feature maps, $F_{final} \in \mathbb{R}^{n_{out}{\times}w{\times}h}$. At the coarsest level, global average pooling (represented by circular arrows in Fig. \ref{fig:psp}) is applied to $F_{final}$ resulting in $n_{out}\times1\times1$ feature maps. Further average pooling operations are also applied that result in $n_{out}\times2\times2$, $n_{out}\times3\times3$ and $n_{out}\times6\times6$ feature map sizes. The final features of the pyramid pooling module are derived by applying a $1\times1$ convolution to each of the resultant feature maps (to ensure equal weighting for each pooling kernel) and upsampling (using bilinear interpolation) to match the dimensions of the final layer feature maps, $F_{psp} \in \mathbb{R}^{n_{psp}{\times}w{\times}h}$. The original final layer feature maps are then concatenated to those derived from the pyramid pooling module ($F_{final} \oplus F_{psp}$) to give a collection of feature maps that capture both local and global context information at varying sub-regions of the input image.

\vspace{-0.1in}
\subsection{Transfer Learning}
We used a pre-trained PSPNet that was trained on natural images from the Pascal VOC dataset \cite{everingham2010pascal}. As the original network architecture accepted 3 input channels for processing RGB images, the network was modified to include an additional \emph{initializer} layer that could accept multi-modal CT perfusion slices. Given a collection of stacked CT perfusion maps, $X \in \mathbb{R}^{5\times256\times256}$ the initializer layer, $L_{init}$, first applies batch normalization \cite{ioffe2015batch} to standardize channel features to a common mean and variance within the batch. Following this, a 1x1 convolutional kernel, $k_{init} \in \mathbb{R}^{5\times3\times1\times1}$ is learned to reduce the channel dimension from 5 to 3. These steps are summarized in Equation (\ref{eqn:linit}), where $BN(\cdot)$ refers to  batch normalization and $*$ refers to the convolution operation.

\begin{equation}
\label{eqn:linit}
L_{init} = BN(X) * k_{init}.
\end{equation}

{\noindent}The resulting feature maps in layer $L_{init} \in \mathbb{R}^{3\times256\times256}$, are ready to be processed using the pre-trained PSPNet weights. Correspondingly, the final layer of the network was modified to replace the 21 class prediction channels, used in Pascal VOC, with binary output channels to predict the presence or absence of ischemic stroke lesions.

Initial fine-tuning took place for all new layers introduced into the network architecture, where pre-trained weights were frozen and
the weights in the newly introduced layers were updated with a learning rate set to $10^{-2}$. Following this, all weights in the network were unfrozen and the network continued training with a learning rate of $10^{-4}$.

\vspace{-0.1in}
\subsection{U-Net and 3D Networks}
We also developed classification models based on the U-Net~\cite{ronneberger2015unet,cciccek20163dunet} and V-Net~\cite{milletari2016vnet} architectures, both commonly used in medical image segmentation. These served as natural comparisons to the PSPNet. The 3D networks also incorporated image depth (axial) information. We used a 2D U-Net and modified versions of the 3D U-Net and V-Net. The modifications reduced the kernel sizes and strides in the $z$ dimension to account for the varying axial depth and minimum depth of 2 in the input images. 

The U-Net architecture contains two pathways, a contracting path which downsamples the image and captures context, and an expanding path which upsamples to perform localization. We modified the base 3D model as described in~\cite{cciccek20163dunet}. In the contracting layer, we reduced the max pooling layer from a size and stride of $2^3$ to $(2\times2\times1)$, down sampling the image in the $x,y$ dimensions by a factor of 2 but leaving the $z$ dimension unchanged. Similarly, in the expanding pathway, we modified the up sampling operation (a transposed convolution) to have a kernel size and stride of $(2\times2\times1)$. For the 2D network, we use the base model as described in~\cite{ronneberger2015unet}. For both models, we modify the input layer for the 5-channel images.

The V-Net model contains a similar contraction and expansion pathway. We modified input to the the base architecture~\cite{milletari2016vnet} to have a kernel size of $3^3$ with unit-padding as the original $5^3$ kernel is too large for our images. Similarly, we modified the convolution and de-convolution (transposed convolution) layers to have kernels with size and stride $(2\times2\times1)$ instead of $2^3$.

\vspace{-0.1in}
\subsection{Loss Function}
We trained the networks using the cross entropy or focal loss~\cite{lin2018focal} functions. Given the true image label for pixel $i$, $y_i \in \{0,1\}$, and a predicted class membership probability $p_i \in [0,1]$, the cross entropy loss is formulated as
\begin{equation}
    CE(p,y) = -y\log(p)-(1-y)\log(1-p),
\label{eqn:cross-entropy}
\end{equation}
and the total loss $\mathcal{L}_{CE}$ is summed over all $N$ pixels,
\begin{equation}
\label{eqn:total-cd}
\mathcal{L}_{CE}=\frac{1}{N}\sum_{i=1}^{N}CE(p_i,y_i).    
\end{equation}
Since our labels are imbalanced, we used a weighted version of the cross entropy loss, 
\begin{equation}
    WCE(p,y) = -wy\log(p)-(1-w)(1-y)\log(1-p),
\label{eqn:wcross-entropy}
\end{equation}
where $w$ is the empirical measure of lesions in the training dataset, $w\in [0,1]$.

The focal loss~\cite{lin2018focal} was introduced as an extension to the cross entropy loss, designed to focus the training on hard to classify examples, by down-weighting easily classified examples, i.e.\ those with high class membership probability. It is formulated as
\begin{equation}
    \label{eqn:focalloss}
    FL(p,y) =  -y(1-p)^{\gamma}\log(p)-(1-y)p^{\gamma}\log(1-p),
\end{equation}
where $\gamma$ is the focusing parameter that governs the down-weighting of the easily classified examples. Note that for $\gamma=0$, the focal loss is the same as the cross entropy loss. With increasing values of $\gamma$, the loss function is smaller for larger values of $p$. Additionally, the function approaches $0$ for smaller values of $p$, allowing the network to focus on the less-confidently classified examples.

\vspace{-0.1in}
\subsection{Implementation details}
RMSProp \cite{Tieleman2012} was used as the optimization routine. The dice coefficient on the validation set was monitored for improvement after every training epoch. If no improvement was observed for 20 epochs, the learning rate was reduced by a factor of 10. A patience flag was set at 50 epochs and if no improvement in the validation dice metric was observed after 50 epochs, early stopping was invoked. For the U-Net and V-Net, the networks were trained from scratch for 200 epochs. The batch size was set to 8 for the 2D networks, and set to 1, using a full image, for the 3D. All models were trained using the PyTorch library on a single Nvidia Titan Xp GPU.

\section{Experiments and Results}
\label{sec:results}
\vspace{-0.1in}
We conducted experiments to determine the optimal network architecture and loss function. The dataset was split into 5 folds. Per-subject folds were created, ensuring no overlap of subjects between folds, i.e.\ subjects with multiple scans existed only within the same fold. We created 5 separate models per architecture, each validated on a distinct fold. We trained different networks using focal loss with $\gamma=1$, and weighted and unweighted cross entropy loss. We evaluated the networks using the Dice Similarity Coefficient (DSC). Given a predicted image label $X$ and the ground truth image label $Y$, the DSC is defined as

\begin{equation}
\label{eqn:dsc}    
DSC(X,Y) = 2\frac{|X \cap Y|}{|X|+|Y|},
\end{equation}
where $|X|$ denotes the cardinality of binary image $X$.

\subsection{5-fold Cross Validation Results}
\begin{figure}
\centering
\includegraphics[width=0.6\columnwidth]{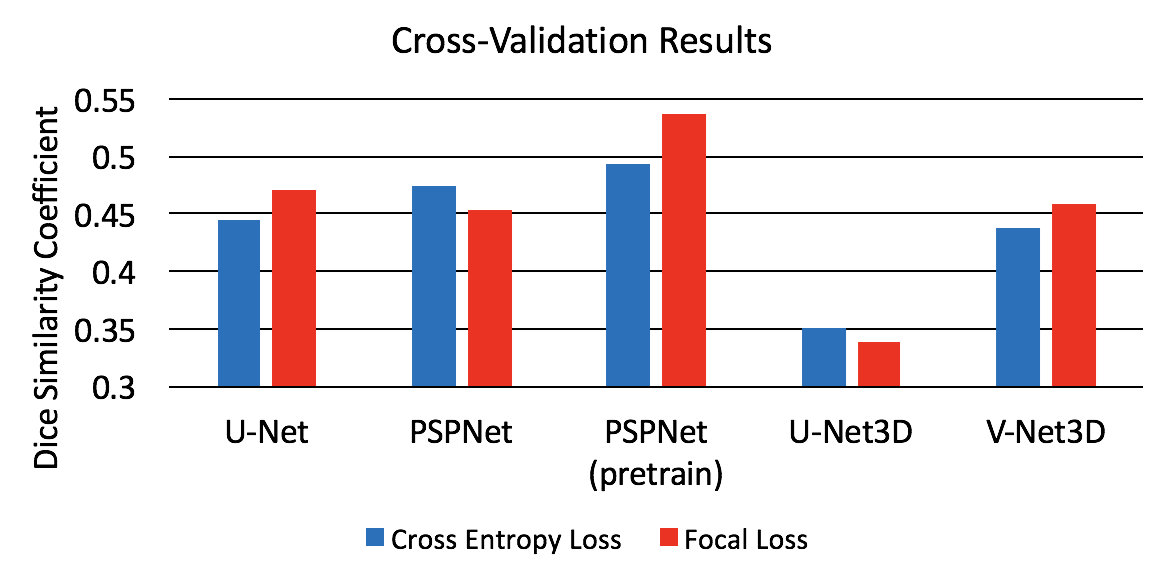}
\caption{5-fold cross validation results on each network architecture, using cross entropy loss (blue) or focal loss (red). The pre-trained PSPNet with focal loss demonstrates the strongest results, with a dice score of 0.54.}
\label{fig:val-results}
\end{figure}

The model parameters for each network were selected based on the best validation dice score. We performed a 5-fold cross-validation to determine model performance. The results are shown on Fig.~\ref{fig:val-results}. The pre-trained PSPNet with focal loss demonstrated the strongest 5-fold cross-validation results ($DSC=0.54$). The 2D U-Net and PSPNet trained from scratch had a similar performance, so the pretraining using our additional input layer improved performance by approximately $7$ dice points. Additionally, the focal loss improved the pre-trained network substantially. Table \ref{tab:folds} shows the per fold DSC results for the pre-trained PSPNet, trained using focal loss and cross entropy. From the table, it can be seen that some validation folds are more challenging than others, leading to varied DSC scores. Overall, usage of focal loss led to an improved overall DSC ($0.54 \pm 0.09$) compared to cross entropy ($0.49 \pm 0.11$). We hypothesize this is due to the fact that the pretraining helps classify the obvious stroke lesion examples, but the focal loss forces the network to learn the more difficult samples.

\begin{table*}
  \centering
    \caption{DSC results per cross validation fold for the PSPNet (pre-trained). Focal loss and cross entropy loss are compared.}
  \begin{tabular}{|p{2cm}|p{4cm}|p{4cm}|}
    \hline
    \bf{Fold} & \bf{Focal Loss} & \bf{Cross Entropy Loss}\\
    \hline
    1 &	0.64	& 0.64\\
    \hline
    2 &	0.42	& 0.37\\
    \hline
    3 &	0.48	& 0.50\\
    \hline
    4 &	0.55	& 0.54\\
    \hline
    5 &	0.58	& 0.41\\
    \hline    
    \bf{Total} &	$\bf{0.54 \pm 0.09}$	& $\bf{0.49 \pm 0.11}$\\
    \hline
  \end{tabular}

  \label{tab:folds}
\end{table*}

\begin{figure}[h]
\centering
\includegraphics[width=0.75\columnwidth]{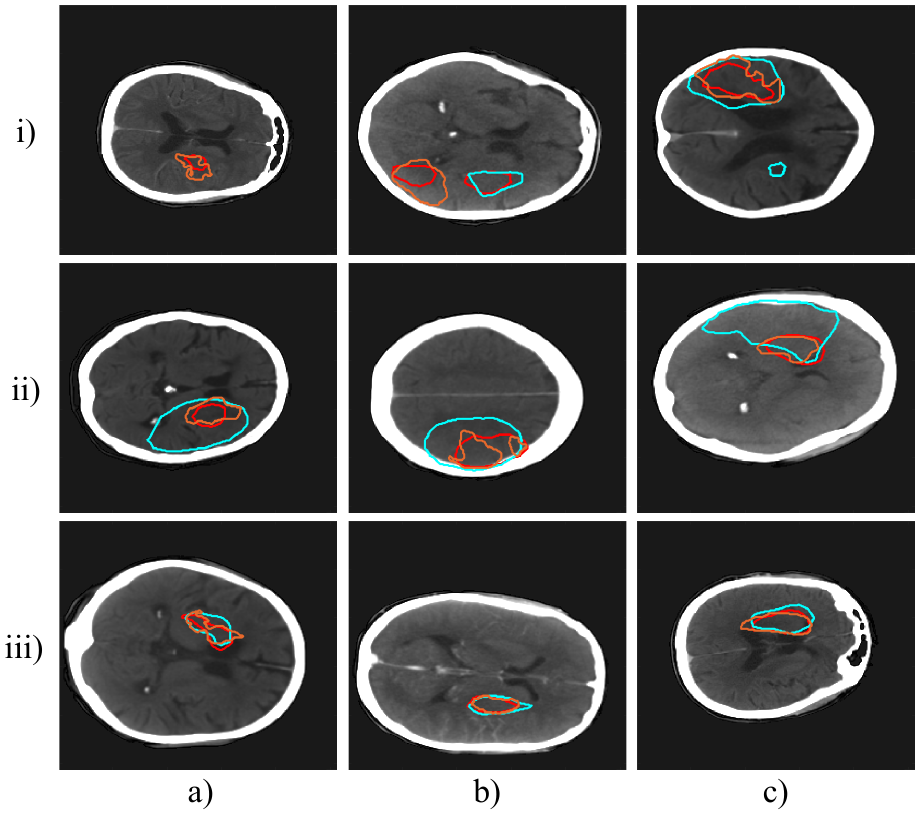}
\caption{Example predicted segmentations on 9 subjects using the pre-trained PSPNet. The ground truth is shown in orange, the network trained with focal loss in red, and the network trained with cross entropy in cyan. In row i), the focal loss network is able to identify difficult lesions and better match the shape of the lesions than the cross entropy network. In the second row, we observe 3 cases where the cross entropy network over-predicts the lesions. Finally, in the the third row, the focal loss network demonstrates closer shape matching than the cross entropy counterpart.  }
\label{fig:result-scans}
\end{figure}

The 3D networks performed poorly. We observed that their increased number of parameters resulted in more overfitting. Additionally, they were unable to take full advantage of the third image dimension, due to the large number of scans with only 2 axial slices. For the two best models, the 2D U-Net and pre-trained PSPNet, we observe the focal loss improved model performance. We demonstrate in Fig.~\ref{fig:result-scans} that the focal loss predicted more fine details in the lesions that were missed by cross entropy in the pre-trained PSPNet. The cross entropy loss network often over-predicted larger lesions than the focal loss network, and the focal loss network was able to more closely predict the fine appearance features of the lesions, and predict areas that cross entropy completely missed.

\vspace{-0.1in}
\subsection{ISLES 2018 Challenge Results}

The challenge evaluated the test data set using the DSC, the Hausdorff Distance (HD), the Average Symmetric Surface Distance (ASSD), precision, recall, and the absolute volume difference (AVD)~\cite{maier2017isles}. The DSC is defined in~\eqref{eqn:dsc}. The HD measures the maximum distance between the two surfaces $X_s$ and $Y_s$, 
\begin{equation}
    HD (X_s,Y_s) = \max\left\{\max_{x\in X_s}\min_{y\in Y_s}d\left(x,y\right), \max_{y\in Y_s}\min_{x\in X_s}d\left(y_s,x_s\right) \right\},
\end{equation}
where $d(\cdot, \cdot)$ is the euclidean distance measure. The ASSD is defined in terms of the average surface distance (ASD),
\begin{equation}
    ASD(X_s,Y_s) = \frac{\sum_{x\in X_s} \min_{y\in Y_s} d(x,y)}{|X_s|},
\end{equation}
and $ASSD = \frac{1}{2}\left(ASD\left(X_s,Y_s\right)+ASD\left(Y_s,X_s\right)\right)$.

Our final submission to the ISLES challenge was an ensemble of ten models that included all five PSPNet (pre-trained) models trained with focal loss, combined with a further five PSPNet (pre-trained) models trained using cross entropy loss. The ensemble achieved a final 5-fold cross validation score of $DSC=0.57$ on the training data leaderboard and $DSC=0.44$ on the testing data leaderboard. The full results of our final model evaluated on the ISLES test set is shown in Table~\ref{tab:results}. The ISLES challenge uses a weighted ranking based on the DSC and Hausdorff Distance to rank submissions. We compare our approach to the performance of the top ranking submission for each metric. Out of a total of 38 submissions to the challenge leaderboard, our approach ranked 6th on DSC, HD and ASSD metrics. Our approach also achieved the second best score on the AVD metric.

\begin{table*}
    \caption{Results of the proposed model compared to the top scores from the ISLES leaderboard, accessed October 2018. Arrows in the header indicate whether lower or higher values are better. $^*$Values normalized by 1,0000,000.}
  \centering
  \begin{tabular}{lcccccc}
    \hline
     & DSC $\uparrow$ & Hausdorff Distance $\downarrow$ & ASSD $\downarrow$ & Precision $\uparrow$ & Recall $\uparrow$ & AVD $\downarrow$\\
    \hline
    Ours &	0.44	&	$1.62^{*}$ &	$1.62^{*}$ & 0.59 & 0.43 & 10.18\\
    Best &	0.51	&	$0.97^{*}$ &	$0.97^{*}$ & 0.62 & 0.58 & 10.08\\
    \bf Place & \pmb{$6^{th}$}/\bf 38 & \bf \pmb{$6^{th}$}/\bf38 & \bf \pmb{$6^{th}$}/\bf38 & \bf \pmb{$3^{rd}$}/\bf 38 & \bf \pmb{$18^{th}$}/\bf 38 & \bf \pmb{$2^{nd}$}/\bf38\\
    \hline    
  \end{tabular}
  \label{tab:results}
\end{table*}

\section{Conclusion}
\vspace{-0.1in}
In this work, we developed fully convolutional neural network models for segmenting ischemic stroke lesions using CTP data. Our model made use of the focal loss function, which demonstrated the ability to identify more fine features in the lesions by focusing on hard to classify examples. We compared models used commonly in medical image segmentation, namely the U-Net and V-Net, with the PSPNet, which was developed for natural image segmentation. 

In future work, we plan to further investigate the role of generative adversarial networks (GANs) in medical image semantic segmentation. In particular GANs can potentially be used to create additional synthetic data for model training. Furthermore, inclusion of a generative loss component within the training procedure could also be investigated. Alternative future work, will also focus on bridging the gap between 2D and 3D CNN models. In general, 3D models did not perform well on the ISLES 2018 dataset, given the limited number of axial slices. Utilizing pre-trained 2D models to better initialize 3D models may hold some potential for improving 3D model performance.

\end{document}